\documentclass[10pt,twocolumn,letterpaper]{article}

\usepackage[pagenumbers]{cvpr} 

\usepackage[dvipsnames]{xcolor}

\usepackage{xspace}
\usepackage{multirow}
\usepackage{color, colortbl}
\usepackage[noend]{algpseudocode}
\usepackage{algorithm}

\usepackage[subtle]{savetrees}

\usepackage{pifont}
\usepackage{xcolor}
\newcommand{\xmarks}{\ding{55}}
\newcommand{\xmark}{{\color{red}\xmarks}}
\newcommand{\cmarks}{\ding{51}}
\newcommand{\cmark}{{\color{ForestGreen}\cmarks}}

\newcommand{\name}{{DriveTrack}\xspace}
\newcommand{\lidar}{{LiDAR}\xspace}

\newcommand{\paragraphb}[1]{\noindent{\bf #1}}

\newcommand{\fig}{{Fig.}\xspace}
\newcommand{\app}{{App.}\xspace}

\newcommand{\davis}{{DAVIS}\xspace}
\newcommand{\kinetics}{{Kinetics}\xspace}
\newcommand{\kubric}{{Kubric}\xspace}
\newcommand{\po}{{PointOdyssey}\xspace}

\newcommand{\tapnet}{{TAP-Net}\xspace}
\newcommand{\tapir}{{TAPIR}\xspace}
\newcommand{\pips}{{PIPs}\xspace}
\newcommand{\pipstwo}{{PIPs++}\xspace}

\newcommand{\nuscenes}{{nuScenes}\xspace}

\definecolor{cvprblue}{rgb}{0.21,0.49,0.74}
\usepackage[pagebackref,breaklinks,colorlinks,citecolor=cvprblue]{hyperref}

\def\paperID{11955} 
\def\confName{CVPR}
\def\confYear{2024}

\title{\name: A Benchmark for Long-Range Point Tracking in Real-World Videos}

\author{
Arjun Balasingam\\
MIT CSAIL
\and
Joseph Chandler\\
MIT CSAIL
\and
Chenning Li\\
MIT CSAIL
\and
Zhoutong Zhang\\
Adobe Systems
\and
Hari Balakrishnan\\
MIT CSAIL
}

\begin{document}
\maketitle

\begin{abstract}

This paper presents DriveTrack, a new benchmark and data generation framework for long-range keypoint tracking in real-world videos. DriveTrack is motivated by the observation that the accuracy of state-of-the-art trackers depends strongly on visual attributes around the selected keypoints, such as texture and lighting. The problem is that these artifacts are especially pronounced in real-world videos, but these trackers are unable to train on such scenes due to a dearth of annotations. DriveTrack bridges this gap by building a framework to automatically annotate point tracks on autonomous driving datasets. We release a dataset consisting of 1 billion point tracks across 24 hours of video, which is seven orders of magnitude greater than prior real-world benchmarks and on par with the scale of synthetic benchmarks. DriveTrack unlocks new use cases for point tracking in real-world videos. First, we show that fine-tuning keypoint trackers on DriveTrack improves accuracy on real-world scenes by up to 7\%. Second, we analyze the sensitivity of trackers to visual artifacts in real scenes and motivate the idea of running assistive keypoint selectors alongside trackers.

\end{abstract}
\section{Introduction}

Long-range keypoint tracking in videos underpins many computer vision applications, including autonomous driving~\cite{avstereo}, robotics~\cite{robotap}, pose estimation~\cite{pose-est}, 3D reconstruction~\cite{locnerf}, and medical imaging~\cite{medicalkeypoint}. Each of these applications involves moving objects and moving cameras. Keypoint tracking---whose goal is to track unique points in the presence of mobility and occlusions---is an active area of research~\cite{tapvid, tapir, pips, pointodyssey}.

Most proposals follow the Track Any Point (TAP)~\cite{tapvid} formulation: given a video and a set of query points, the algorithm must estimate the locations of those points in all other frames where they are visible. The underlying tracking algorithms vary significantly. TAPIR~\cite{tapvid, tapir} is an end-to-end method that predicts correspondences using feature maps and cost volumes. By contrast, \pipstwo~\cite{pips, pointodyssey} stitches optical flow vectors together to construct long-range trajectories. These are two recent methods that improve the state-of-the-art, adding to a number of techniques proposed over the last two decades.

\begin{figure}[t]
    \centering
    \includegraphics[scale=0.65]{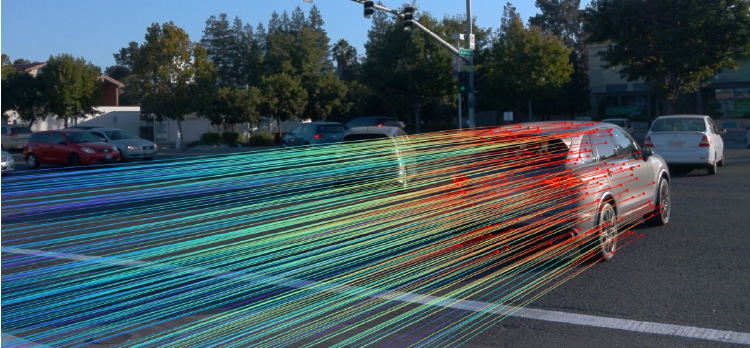}
    \vspace{-5pt}
    \caption{\name automatically generates dense, accurate, and long-range point track annotations for autonomous driving videos.}
    \vspace{-10pt}
    \label{fig:intro:teaser}
\end{figure}

This paper observes that the accuracy of state-of-the-art trackers suffers on real videos. In particular, noisy visual characteristics---such as texture, lighting variations, occlusions, and motion-induced image distortions---can hinder tracking performance (\S\ref{sec:motivation}). The key problem is that modern trackers train on vast synthetic datasets~\cite{kubric, pointodyssey, rgbstacking}, whose scenes do not exhibit these imperfections. There exist only two benchmarks for the TAP task on real-world videos~\cite{kinetics, davis}, with each offering tens of human-labeled annotations per scene.

\begin{table*}
    \footnotesize
    \centering
    \begin{tabular}{r|cccccc}
        \hline
         &  FlyingThings++~\cite{pips} &  \kubric~\cite{kubric}& \kinetics~\cite{kinetics} &\davis~\cite{davis} &\po~\cite{pointodyssey} &\textbf{\name}\\ \hline \hline        
         Resolution& $540 \times 960$ & $256 \times 256$ & $\geq 720 \times 1080$ & $1080 \times 1920$ & $540 \times 960$ & $1280 \times 1920$ \\
         \rowcolor{gray!20}
         Frame rate (Hz)& 8 & 8 & 25 & 25 & 30 & 10 \\
         Avg. trajectory count& 1024 & Arbitrary & 26.3 & 21.7 & 18,700 & 100,800 \\
         \rowcolor{gray!20}
         Avg. frames per video& 8 & 24 & 250 & 67 & 2,035 & 84 \\ \hline
         \# of training frames& 21,818 & Arbitrary & -- & -- & 166,000 & 672,000 \\
         \rowcolor{gray!20}
         \# of validation frames& 4,248 & Arbitrary & -- & -- & 24,000 & 84,000 \\
         \# of test frames& 2,247 & Arbitrary & 297,000 & 1,999 & 26,000 & 84,000 \\
         \rowcolor{gray!20}
         \# of point annotations& 300M & Arbitrary & 80M & 400K & 49B & 84B \\
         \# of point tracks& 16K & Arbitrary & 32K & 650 & 11K & 1B \\ \hline
         \rowcolor{gray!20}
         Real-world scenes& \xmark & \xmark & \cmark & \cmark & \xmark & \cmark \\
         Depth maps& \cmark & \cmark & \xmark & \xmark & \cmark & \cmark\\
         \rowcolor{gray!20}
         Object masks& \cmark & \cmark & \cmark & \cmark & \cmark & \cmark \\
         Multiple views& \xmark & \cmark & \xmark & \xmark & \cmark & \cmark \\ \hline
    \end{tabular}
    \vspace{-5pt}
    \caption{\name is the first benchmark on real-world videos to offer annotations that match the scale and fidelity of synthetic datasets.}
    \vspace{-10pt}
    \label{tab:overview}
\end{table*}

To overcome this shortcoming, we propose {\em\name}, a large-scale benchmark for long-range point tracking tasks. \name brings to real-world videos the density and fidelity of annotations available only for synthetic benchmarks today. By using camera feeds from cars driven in urban areas, \name captures realistic motion, noisy visual attributes, and occlusions, which synthetic~\cite{kubric, rgbstacking} or rendered~\cite{pointodyssey} datasets do not model. \fig~\ref{fig:intro:teaser} visualizes the annotations computed by \name for a driving scene. Although \name is built on autonomous driving videos, it captures the wide variety of visual artifacts typical in real-world scenes.

To generate point tracks, we adapt methods used by synthetic benchmarks~\cite{kubric, pointodyssey} that use rendering software to precisely annotate the motion of simulated trajectories. However, real-world videos do not have the luxury of a simulator. To overcome this challenge, \name leverages timestamped \lidar point clouds, object bounding box annotations, and camera poses and orientations~\cite{waymo, nuscenes, kitti, lyft}. Since \lidar point sweeps do not have 1:1 correspondence over time~\cite{waymo}, \name cannot compute correspondences between adjacent point clouds, as synthetic benchmarks are able to. \name instead transforms each timestamped point cloud according to the camera pose and bounding box annotations to generate hundreds of thousands of highly accurate point annotations per object. We also implement several refinements to ensure that our annotations are robust to noise in hand-labeled bounding boxes.

\name can annotate point tracks for rigid bodies in any dataset of real-world videos that includes point clouds, 3D object segmentations, and camera poses. With this paper, we release point tracking annotations for the Waymo Open Dataset~\cite{waymo}. Our dataset contains 1 billion point tracks for over 10,000 distinct objects across 24 hours of video. Table~\ref{tab:overview} compares \name to other point tracking benchmarks.

\name makes long-range tracking practical for real-world scenes. This paper presents results for two use cases:
\begin{itemize}
    \item \textbf{Fine-tuning keypoint trackers}. We fine-tune \tapnet~\cite{tapvid}, \tapir~\cite{tapir}, \pips~\cite{pips}, and \pipstwo~\cite{pointodyssey} on \name, showing an improvement of 4-7\% on \name's test set and 1-2\% on \davis~\cite{davis} (\S\ref{sec:finetune}).
    
    \item \textbf{Keypoint sensitivity.} Visual imperfections make keypoint tracking in real videos more challenging. We use the scale of annotations made available by \name to quantify the sensitivity of tracking accuracy to visual imperfections. From this analysis, we motivate how \name can be used to build keypoint selectors, which can recommend robust keypoints to use with trackers (\S\ref{sec:sensitivity}).
\end{itemize}
\name's data generation code and point tracking annotations on the Waymo dataset are available at \url{drivetrack.csail.mit.edu}.

\section{Related Work}

\paragraphb{Real-world datasets.}
The TAP-Vid benchmark~\cite{tapvid} released annotations for DAVIS~\cite{davis} and Kinetics~\cite{kinetics}, two real-world video datasets. TAP-Vid introduces a workflow that helps a human annotate keypoint tracks in a sequence of video frames. State-of-the-art keypoint trackers~\cite{pips, tapir, tapvid} evaluate on the \davis and \kinetics benchmarks to quantify performance on real-world scenes. However, since the scale of annotations is on the order of tens of trajectories per scene, these benchmarks are viable only for evaluation, and not for fine-tuning models to be robust to visual artifacts seen in real-world scenes.

The KITTI dataset's Segmenting and Tracking Every Pixel (STEP) benchmark~\cite{kitti} tags every pixel (or 3D point) with a semantic label (\eg car, truck, pedestrian, \etc) and a unique track ID. It constructs tracks by propagating segmentation masks using RAFT~\cite{raft}. However, STEP does not annotate point correspondences over a long-range, \eg the exact trajectory of a unique pixel over a video sequence. 
Even earlier, the Middlebury dataset~\cite{middlebury} was the de-facto benchmark for optical flow and motion estimation tasks. This dataset consists of a mix of real-world and synthetic scenes. While the annotation quality is high, the sheer number of annotations, as with \davis and \kinetics, is small.

\paragraphb{Synthetic benchmarks.}
Most of the data used to train and evaluate long-range motion estimation has been synthetic. Popular benchmarks include FlyingChairs~\cite{flyingchairs}, FlyingThings3D~\cite{flyingthings3d}, and AutoFlow~\cite{autoflow} providing short-range (\ie 2 frames) annotations, and \kubric~\cite{kubric}, \po~\cite{pointodyssey}, and FlyingThings++~\cite{pips} providing long-range labels. These datasets contain different variants of generated objects moving in random directions on random backgrounds. Of these benchmarks, \po has annotations with the greatest volume and fidelity, and uses rendering tools and real motion traces to synthesize photo-realistic scenes. However, it fails to capture visual artifacts characteristic of real scenes that hamper the performance of keypoint trackers (\S\ref{sec:motivation}).

\paragraphb{Keypoint tracking.}
Datasets with dense and accurate point tracks are critical to developing and evaluating keypoint tracking methods that follow the TAP formulation. \tapir~\cite{tapir}, \tapnet~\cite{tapvid}, \pips~\cite{pips}, and PIPs++~\cite{pointodyssey} are four methods that have pushed the state-of-the-art over the last two years. Other approaches to keypoint tracking rely on optical flow~\cite{raft} or on structure-from-motion~\cite{particlesfm}.
The focus of this paper is on \name and the new use cases it enables: fine-tuning trackers on real-world scenes and an analysis of keypoint sensitivity.
\section{Motivation}
\label{sec:motivation}

Keypoint tracking on videos is a decades-old problem in computer vision, but it is also evolving rapidly. Over the last two years, four keypoint trackers~\cite{tapir, tapvid, pips, pointodyssey} have pushed state-of-the-art results. All recent methods train on large-scale synthetic benchmarks~\cite{kubric, flyingchairs, flyingthings3d, pointodyssey}. In this section, we make the case for a similar point tracking benchmark for real-world scenes by highlighting current limitations in the performance of state-of-the-art keypoint trackers.

\begin{figure}[t]
    \centering
    \includegraphics[scale=0.65]{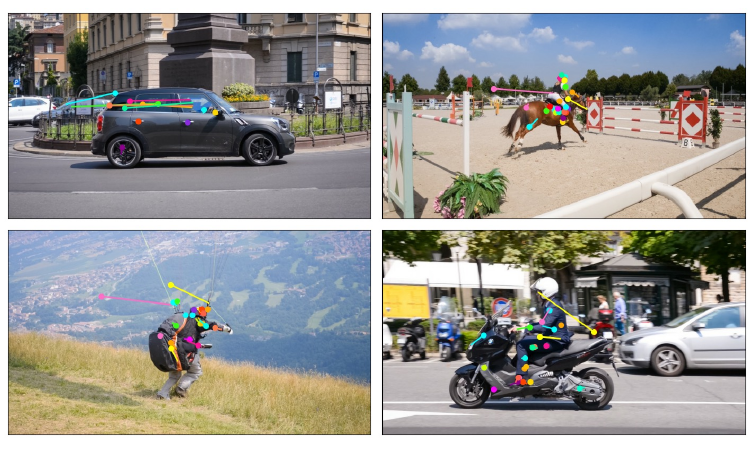}
    \vspace{-5pt}
    \caption{State-of-the-art keypoint trackers struggle on real-world scenes with complex lighting and texture attributes. Shown here are point tracks predicted by TAPIR~\cite{tapir} relative to their ground-truth locations for 4 scenes from the DAVIS dataset~\cite{davis}.}
    \vspace{-10pt}
    \label{fig:motivation:errors}
\end{figure}

\paragraphb{Keypoint trackers suffer on real videos.}
\fig~\ref{fig:motivation:errors} shows the tracking error achieved by \tapir~\cite{tapir} on four scenes from the \davis dataset~\cite{davis}. The markers indicate the locations predicted by \tapir, and the line segments lead to their respective ground-truth locations.

Notice that the tracking quality varies significantly. For instance, \tapir tracks the car turning the roundabout (top left) accurately around well-defined corners and edges, like the door handles and wheel spokes. However, the predicted tracks drift  from ground truth on reflective surfaces like the windows. In the paragliding video (bottom left), \tapir struggles on patches of the parachute with high-frequency lighting content. Tracking accuracy is also poor at the suspension ropes, which blend with the saturated background.

\paragraphb{Limitations of synthetic datasets.}
While the synthetic benchmarks on which these models are trained offer annotations at a large scale and fidelity, they do not exhibit visual artifacts that are all too common in real-world scenes. Synthetic benchmarks, such as \kubric~\cite{kubric}, FlyingThings++~\cite{flyingthings3d}, and \po~\cite{pointodyssey}, only capture rudimentary lighting conditions like shadows and selects from a corpus of simplistic rendered objects. Textures are simple and lighting patterns are monotonic. Given these limitations, it is unrealistic to expect these keypoint trackers to excel on real-world scenes.

\paragraphb{Annotating real-world datasets.}
The problem is that there do not exist any benchmarks on real-world scenes that offer the fidelity and scale of annotations available for synthetic benchmarks. Existing benchmarks on real-world scenes, like \davis and \kinetics, curate only a handful of human-labeled keypoint tracks per scene. Moreover, the annotated keypoints are biased toward locations that are naturally easier for a human to track.

A key requirement for annotating datasets accurately and automatically is 3D context. Autonomous driving datasets include 3D bounding boxes and \lidar point clouds. However, annotating in the presence of these sparse and noisy labels introduces several challenges, for which we propose solutions in \S\ref{sec:data}.

\section{\name Overview}
\label{sec:data}

\begin{figure}[t]
    \centering
    \includegraphics[scale=0.65]{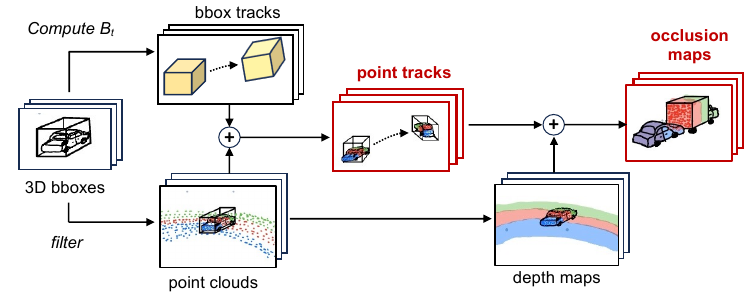}
    \vspace{-5pt}
    \caption{\name transforms each timestamped point cloud according to the vehicle's camera poses and a target object's bounding boxes to automatically and accurately annotate point tracks for that object.}
    \label{fig:data:overview}
\end{figure}

\begin{table}[t]
    \footnotesize
    \centering
    \begin{tabular}{c|p{6cm}}
         \textbf{Notation}& \textbf{Meaning}\\ \hline
         $N$& number of points tracked \\
         \rowcolor{gray!20}
         $F$& duration of RGB video \\
         $R$& camera extrinsic matrix\\
         \rowcolor{gray!20}
         $K$& camera intrinsic matrix\\
         $W_t$& transforms point from vehicle frame to world frame at time $t$\\
         \rowcolor{gray!20}
         $B_t$& transforms bounding box from origin to location in world frame at time $t$\\
         $X_t$ & target object's \lidar point cloud at time $t$ \\
         \rowcolor{gray!20}
         $\hat{X}^{(t)}_\tau$ & projection of $X_t$ at time $\tau$ \\
         $T_V$ & set of 3D point tracks in vehicle frame \\
         \rowcolor{gray!20}
         $\hat{T}_I$ & set of 2D point tracks in image space \\
         $\hat{D}$ & depth map \\
         \rowcolor{gray!20}
         $\hat{O}$ & occlusion map \\
    \end{tabular}
    \vspace{-5pt}
    \caption{Notation for \name's point tracking workflow.}
    \vspace{-10pt}
    \label{tab:notation}
\end{table}

\name leverages autonomous driving datasets~\cite{waymo, nuscenes, kitti} to generate dense point tracks on real-world videos. \fig~\ref{fig:data:overview} illustrates \name's data generation workflow, which combines \lidar sweeps, 3D bounding box annotations, sensor calibrations, and camera poses/orientations to derive keypoint annotations with high fidelity.

\name computes point tracks by shifting recorded point clouds by the poses of the driving (ego) vehicle. To create the benchmark, we aggregate and filter the timestamped data (\S\ref{sec:data:prep}), transform and track the point clouds (\S\ref{sec:data:track}), and estimate occlusions (\S\ref{sec:data:occ}) by computing depth maps (\S\ref{sec:data:depth}). We release annotations on the Waymo Open Dataset~\cite{waymo}, but our data generation workflow is compatible with other autonomous driving datasets (\S\ref{sec:data:robustness}). Table~\ref{tab:notation} introduces notation that we use in the rest of this section to formalize the point tracking procedure. \name computes a set of 2D point tracks $\hat{T}_I \in \mathbb{R}^{N \times F \times 2}$, as well as an occlusion map $\hat{O} \in \mathbb{R}^{N \times F}$ that indicates when points are not visible from the perspective of the camera.

\subsection{Dataset preparation and requirements}
\label{sec:data:prep}

Autonomous driving datasets split data by different modalities. \name pre-processes this data by joining tables by timestamp, and bundles the following for each unique object $k$:
\begin{itemize}
    \item We extract a length-$F$ sequence of RGB frames containing the object, along with camera calibration matrices $R$ and $K$. We consider each of the cameras on the ego vehicle separately.
    \item We convert the bounding box annotations for each unique object $k$ into a sequence of timestamped transformations $B_{k,t}$ that describe how to transform the object from the world origin to its position in the global frame.
    \item We extract the \lidar point clouds $X_{k,t}$ by filtering the scene's point cloud at time $t$ by object $k$'s bounding box.
    \item We export transformations $W_t$ that map points in the ego vehicle's reference frame to the world frame.
\end{itemize}
For simplicity, we omit the object index $k$ in the rest of this paper.

\subsection{Point tracking}
\label{sec:data:track}

\name computes correspondences in 3D and then projects those points to the image space. We compute point tracks separately for each target object that is annotated in the dataset.

\paragraphb{Tracking with point clouds.}
Unlike synthetic benchmarks, \name cannot render the scene in a simulator and generate infinitely many points to track. It instead is confined to a finite set of points sampled by the ego vehicle's \lidar sensor. However, since \lidar samples different points in each sweep, the point clouds in adjacent timesteps $X_t$ and $X_{t+1}$ will not have a 1:1 correspondence.

\name instead takes each point cloud $X_t$ and ``shifts'' it across time according to the bounding annotations, constructing a sequence of point clouds $\{ \hat{X}_{\tau}^{(t)} \forall \tau \in [1, F] \}$.  
To project $X_t$, we need to account for the motion of the target object and the motion of the ego vehicle. The transformation $B_t$ captures the location of the object at each time $t$ in the global frame, and $W_t$ specifies how to transform a point from the ego vehicle's reference frame to the global reference frame at each time $t$. Thus, to project the point cloud, we (i) project to the global frame, (ii) transform to bounding box coordinate system, (iii) apply the bounding box transformation in the next timestep, and (iv) project back to the vehicle's reference frame. Formally, the projected point cloud at $\tau$ is:
\begin{equation}
    \hat{X}_{\tau}^{(t)} = W_{\tau}^{-1} B_{\tau} B_{t}^{-1} W_{t} X_{t}
\end{equation}
This yields $N$ length-$F$ tracks, where $N$ is the number points in $X_t$.\footnote{This assumes (for simplicity) that each point cloud $X_t$ has $N$ points.} We can repeat this procedure for {\em each} $t \in [1, F]$. In total, this gives $N \times F$ point tracks of length $F$ for this target object.

\paragraphb{Point tracks.}
We can define the matrix $\hat{T}_V \in \mathbb{R}^{N \times F \times 3}$ to hold the point tracks projected using the procedure described above. Formally, if $\hat{T}_V^{(i)} = \left[ \hat{X}_1^{(i)} \dots \hat{X}_t^{(i)} \dots \hat{X}_F^{(i)} \right]$, then $\hat{T}_V = \left[ \hat{T}_V^{(1)^\intercal} \dots \hat{T}_V^{(i)^\intercal} \dots \hat{T}_V^{(F)^\intercal} \right]$. We can then project these points to the image space using the camera matrices as follows: $\hat{T}_I = K R^{-1} \hat{T}_V$.

\subsection{Occlusion estimation}
\label{sec:data:occ}

\name flags a point as occluded if its 3D position is further from the camera than the nearest physical point in the same direction~\cite{kubric, pointodyssey}. If $\hat{x}_t \in \mathbb{R}^3$ is the 3D projection of a point $p \in \mathbb{R}^2$ relative to the ego vehicle's reference frame, then $\hat{d}_t = \| \hat{x}_t \|$ gives the distance of that point from the camera. To find the nearest physical points from the camera, we compute depth maps $\hat{D}_t \in \mathbb{R}^{W \times H}$.\footnote{\S\ref{sec:data:depth} describes how we compute depth maps from the point clouds.} Then, a point $p = (p_x, p_y)$ is occluded if $\hat{d}_t > \hat{D}_t[p_x, p_y]$. We repeat this for all $p \in \hat{T}_I$ to compute $\hat{O}$.

\begin{figure}
    \centering
    \includegraphics[scale=0.35]{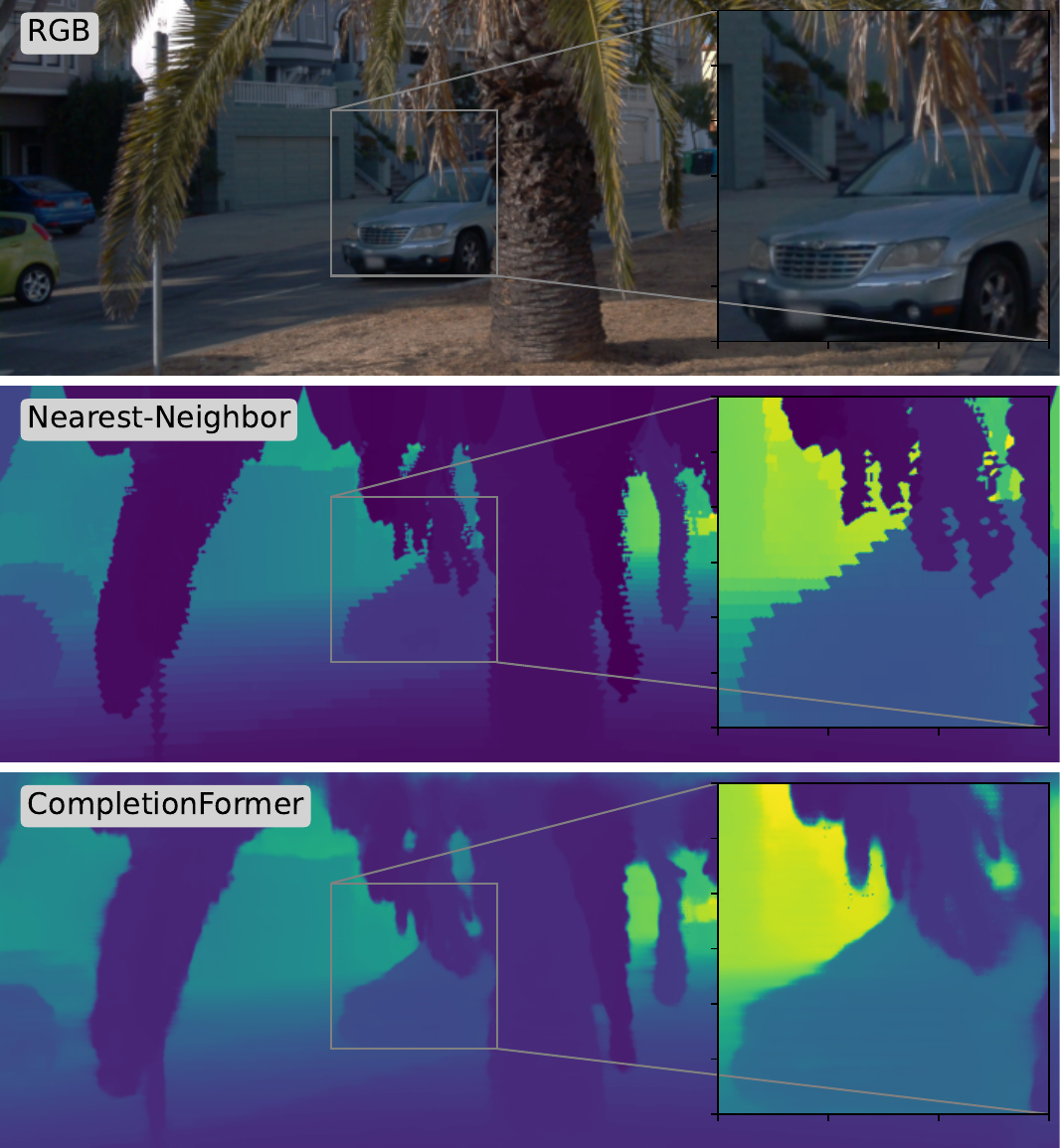}
    \vspace{-5pt}
    \caption{Depth maps computed by Nearest Neighbor and CompletionFormer~\cite{completionformer} for a scene in the Waymo dataset~\cite{waymo}.}
    \vspace{-10pt}
    \label{fig:data:depth}
\end{figure}

\subsection{Depth completion}
\label{sec:data:depth}

Synthetic benchmarks export dense depth maps. However, the autonomous driving datasets that \name uses for its point tracking annotations only export {\em sparse} \lidar point clouds, \ie a collection of 3D points that the sensor samples through several sweeps. 
To create dense depth maps, \name first projects the 3D point cloud to 2D, creating a sparse depth map. Then, it leverages depth completion methods that learn a function $f$ to interpolate a set of 2D points to a dense 3D map. We denote $X_{I,t}$ as the 2D projection of the ground-truth point cloud $X_t$ at time $t$. $f$ maps $X_{I,t}$ to a dense depth map $\hat{D}_{t}$, which we then use to estimate occlusions.

We consider two depth completion models. The first interpolates nearest neighbors in the image plane, by assigning each pixel in the image the same depth value as the closest ground-truth sparse depth point in $X_{I,t}$ by 2D Euclidean distance.
The second method is CompletionFormer~\cite{completionformer}, a deep depth completion model that achieves state-of-the-art performance on the Kitti Depth Completion (DC) benchmark~\cite{kittidepth}. CompletionFormer uses both the sparse depth map and the RGB source image to produce a dense depth map. It uses non-local spatial propagation networks~\cite{nlspn} to share affinity information from depth areas in the image about which the model has high confidence to areas with lower confidence. For both models, we interpolate depths at floating point pixel values~\cite{kubric,pointodyssey}, by computing a max pooling over the neighborhood of the four corner pixels around which we interpolate. Max pooling overestimates depth values compared to an alternative like bilinear interpolation, but we find that it yields more accurate occlusion maps.

\fig~\ref{fig:data:depth} visualizes depth maps computed by Nearest Neighbor (middle) and CompletionFormer (bottom) for a scene. 
Empirically, we find that Nearest Neighbor produces accurate depth and occlusion maps on our benchmark. However, it produces jagged artifacts around the edges of objects, particularly in the foreground of the scene. Consequently, occlusion estimates tend to be inaccurate on the boundaries of vehicles or on other sharp gradients in depth.
By contrast, CompletionFormer handles object edges more gracefully than the nearest-neighbor interpolation, by leveraging non-local spatial propagation.
However, it suffers particularly on thin background objects like tree leaves or signposts. 
\S\ref{sec:finetune} reports results from fine-tuning state-of-the-art trackers on versions of \name built using both Nearest Neighbor and CompletionFormer~\cite{completionformer}.

\subsection{Robustness}
\label{sec:data:robustness}

\paragraphb{Filters.}
We find that noise from the hand-labeled bounding box annotations and interpolation errors from depth completion methods can degrade track quality. We implement several filters to ensure that the computed point tracks are accurate:
\begin{itemize}
    \item Object must lie in the camera's field-of-view (either visible or occluded) for at least 24 frames. The exact duration does not matter; we choose 24 frames because it is the video length used to train \tapir~\cite{tapir} and \tapnet~\cite{tapvid}. This ensures that videos are not too short for model training and are sufficient length to learn robust point tracks.
    \item Object must be within 20m of the camera at some point in the video sequence. We find that the quality of depth completion is inferior and highly variable for more distant objects, impacting \name's ability to determine occlusions accurately.
    \item Object must lie in the camera's field-of-view continuously for at least 24 frames. Moreover, the object cannot fully leave the frame and return for simplicity, although we could allow this in future iterations.
\end{itemize}
Additionally, we only use the front, front-left, and front-right cameras. The side cameras often have short object tracks. 

\paragraphb{Other autonomous driving datasets.}
We implement the workflow described above for the Waymo Open Dataset~\cite{waymo}. However, \name's data generation workflow is compatible with the heavily-curated domain of autonomous driving videos~\cite{kitti, nuscenes, lyft}. 
Porting \name to other datasets might require more careful attention to interpolation, which we leave to future work. For instance, the \nuscenes dataset~\cite{nuscenes} has annotations at 2 Hz, and requires interpolating bounding boxes to match the 10 fps video frame rate. Moreover, \nuscenes's dataset has sparser point clouds, which would yield fewer total point tracks and poorer depth maps. We leave additional refinements to future work.

\paragraphb{Implementation.}
App.~\ref{app:impl} discusses optimizations that we added to \name to improve annotation speed.

\section{\name Analysis}

\begin{figure*}[t]
    \centering
    \includegraphics[width=\textwidth]{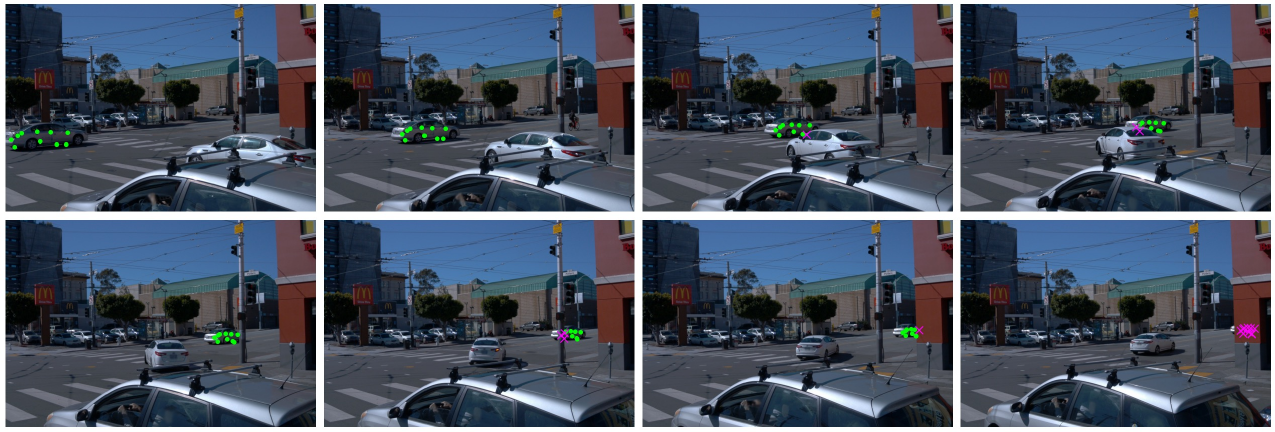}
    \vspace{-15pt}
    \caption{Visible point correspondences ({\color{green}$\bullet$}) and occlusions ({\color{magenta}$\times$}) computed by \name over eight samples from a 30-frame sequence.}
    \label{fig:waymo-strip}
\end{figure*}

\paragraphb{Point track quality.}
Synthetic benchmarks~\cite{kubric, pointodyssey} generate exact point tracks by using rendering tools to simulate motion. However, as we note in \S\ref{sec:data:robustness}, \name is susceptible to annotation error. \fig~\ref{fig:waymo-strip} shows, for a representative scene in the Waymo dataset~\cite{waymo}, point correspondences computed by \name. We denote visible points with $\bullet$ and occluded points with $\times$. Note that the point tracks are extremely accurate. Our website\footnote{\url{drivetrack.csail.mit.edu}} includes several videos that illustrate the quality of \name's annotations. App.~\ref{app:annotations} includes more annotated examples, including an edge case where \name fails.

\begin{table}
    \scriptsize
    \centering
    \begin{tabular}{lccccc}
        \hline
         \multirow{2}{*}{Depth Completion}&\multicolumn{4}{c}{Speed Error (m/s)} \\
         & p25 & p50 & p75 & p95 & p99 \\ \hline
         Nearest Neighbor & 0.0651 & 0.1260 & 0.2471 & 0.5497 & 0.8687 \\
 CompletionFormer~\cite{completionformer} & 0.0702 & 0.1273 & 0.2663 & 0.4874 & 0.6615 \\ \hline
    \end{tabular}
    \vspace{-5pt}
    \caption{The speed of \name's point tracks is consistent (\ie low error) with the speed of Waymo's ground-truth bounding boxes~\cite{waymo}.}
    \vspace{-10pt}
    \label{tab:speed}
\end{table}

\paragraphb{Point track consistency.}
Since we do not have reference ground truth tracks, we instead compare the velocity of the computed point tracks against the labeled velocities for object annotations, as a proxy for annotation error. Since \name tracks rigid bodies, we expect the velocity for each point track for an object to match the annotated velocity for that object. To quantify consistency, we compute the average velocity of a point track and compare the distribution (\ie 25-99pct) of those point velocities to the average annotated velocity for that object. Table~\ref{tab:speed} shows the median value of different percentiles across all objects in the dataset. We report results for both depth completion methods (\S\ref{sec:data:depth}). The median error in speed is 0.13 m/s, indicating that \name is faithful to Waymo's ground-truth annotations.

\paragraphb{Statistics.}
The \name benchmark on the Waymo dataset~\cite{waymo} includes 1000 scenes across 3 different cameras, totaling about 10,000 $1280 \times 1920$ videos at 10 fps. Each video is for a single object with 100,000 trajectories on average per video. We further split our dataset into 800 scenes for training, 100 scenes for validation, and 100 scenes for test, which translates to about 8,000, 1,000, and 1,000 videos respectively, with an average video length of 84 frames. Table \ref{tab:overview} compares \name with other standard datasets for point tracking.

\section{Fine-tuning keypoint trackers}
\label{sec:finetune}

\begin{table*}
    \centering
    \footnotesize
    \begin{tabular}{ll|ccc|ccc|ccc} \hline
         \multirow{2}{*}{Tracker} & \multirow{2}{*}{Training} & \multicolumn{3}{c|}{Kubric~\cite{kubric}}&\multicolumn{3}{c|}{\davis~\cite{davis}} &\multicolumn{3}{c}{\name} \\ 
         & & AJ& $< \delta^{x}_{avg}$&OA&AJ& $< \delta^{x}_{avg}$&OA& AJ& $< \delta^{x}_{avg}$&OA \\
         \hline
         \hline
         \multirow{2}{*}{\tapnet~\cite{tapvid}} & \kubric~\cite{kubric} & \textbf{65.4} & \textbf{77.7} & \textbf{93.0} & 38.4 & 53.1 & \textbf{82.3} & 63.6 & 73.8 & 92.4 \\
         & + \name & 37.0 & 54.0 & 83.5 & \textbf{39.2} & \textbf{54.7} & 78.6 & \textbf{70.3} & \textbf{80.4} & \textbf{93.2} \\
         \hline
         \multirow{2}{*}{\tapir~\cite{tapir}} & Panning \kubric & \textbf{84.7} & \textbf{92.1} & \textbf{95.8} & 62.8 & 74.7 & \textbf{89.5} & 78.8 & 87.1 & 94.4 \\ 
          & + \name & 80.8 & 89.6 & 93.8 & \textbf{64.0} & \textbf{76.1} & 88.0 & \textbf{84.1} & \textbf{90.9} & \textbf{95.1} \\
         \hline
         \multirow{2}{*}{\pipstwo~\cite{pointodyssey}} & \po~\cite{pointodyssey} & -- & \textbf{25.8} & -- & -- & 70.4 & -- & -- & 81.5 & -- \\
          & + \name & -- & 25.6 & -- & -- & \textbf{71.2} & -- & -- & \textbf{85.3} & --\\
         \hline
    \end{tabular}
    \vspace{-5pt}
    \caption{Tracking performance of different models on different datasets before/after fine-tuning on \name. $\delta_{avg}^{x}$ measures positional tracking accuracy, OA measures binary occlusion accuracy, and AJ considers both position and occlusions. We report all metrics as percentages. Higher is better.}
    \vspace{-10pt}
    \label{tab:finetuning}
\end{table*}

\begin{table}[t]
    \centering
    \footnotesize
    \begin{tabular}{lccc}
    \hline
    \multirow{2}{*}{Depth Completion}& \multicolumn{2}{c}{Average Jaccard (AJ)}\\
         &  \davis~\cite{davis} & \name\\ \hline \hline
         Nearest Neighbor& \textbf{39.29} & \textbf{70.3} \\
         CompletionFormer~\cite{completionformer}& 38.60 & 69.5 \\ \hline
    \end{tabular}
    \vspace{-5pt}
    \caption{Comparing the Nearest Neighbor and CompletionFormer depth completion strategies for \name on fine-tuning \tapnet.}
    \vspace{-10pt}
    \label{tab:finetune-depth}
\end{table}

\name is large enough that, for the first time, we can train keypoint trackers on real-world scenes. In this section, we show results from fine-tuning three state-of-the-art keypoint trackers~\cite{tapvid, tapir, pointodyssey} on \name.

\subsection{Setup and metrics}

\paragraphb{Dataset.}
We work with a subset of annotations from \name, comprising of 300 target objects split across 100 different scenes. We split each annotation into 24-frame subsets, resulting in 900 distinct training examples, 22,000 training frames, and 168 million point trajectories. We use the Nearest Neighbor depth completion model for most experiments.

\paragraphb{Experimental setup.}
We use the official code for \pipstwo~\cite{pointodyssey}, \tapnet~\cite{tapvid}, and \tapir \cite{tapir}, and the pre-trained weights released with each paper. For \tapir, we use the panning Kubric model checkpoint as described in their paper, as opposed to the one trained on the original MOVi-E dataset. We train all models for 5000 steps on 8 V100 GPUs with 500 warmup steps. For \tapnet and \tapir, we use a learning rate of $1 \times 10^{-5}$ and an AdamW optimizer with $\beta_1 = 0.9$ and $\beta_2 = 0.95$ and weight decay $1 \times 10^{-2}$. For \pipstwo, we use a learning rate of $1 \times 10^{-5}$ and a AdamW optimizer with the default $\beta$ parameters and a weight decay of $1 \times 10^{-6}$. For all models, we halt fine-tuning when performance on a given task has reached a peak.

\paragraphb{Metrics.}
We evaluate these models on three standard TAP-Vid metrics~\cite{tapvid}. $< \delta^{x}$ evaluates the positional accuracy for visible points, by measuring the fraction of points that are within a threshold of their ground-truth locations. We report $< \delta^{x}_{avg}$, which averages across 5 thresholds: 1, 2, 4, 8, and 16 pixels. Occlusion Accuracy (OA) reports the classification accuracy for the occlusion status predicted for each point. Average Jaccard (AJ) measures both position and occlusion accuracy. Jaccard is the fraction of ``true positives'' (\ie points within the threshold of visible ground-truth points) divided by ``true positives'' plus ``false positives'' (\ie points predicted visible, when the ground-truth reports occluded or farther than threshold) plus ``false negatives'' (\ie ground-truth visible points that are predicted as occluded or farther than the threshold). AJ averages Jaccard across the same thresholds as $< \delta^{x}_{avg}$. Note that we do not report AJ and OA for \pipstwo since it does not export occlusions~\cite{pointodyssey}.

\begin{figure*}[t]
    \centering
    \includegraphics[width=\textwidth]{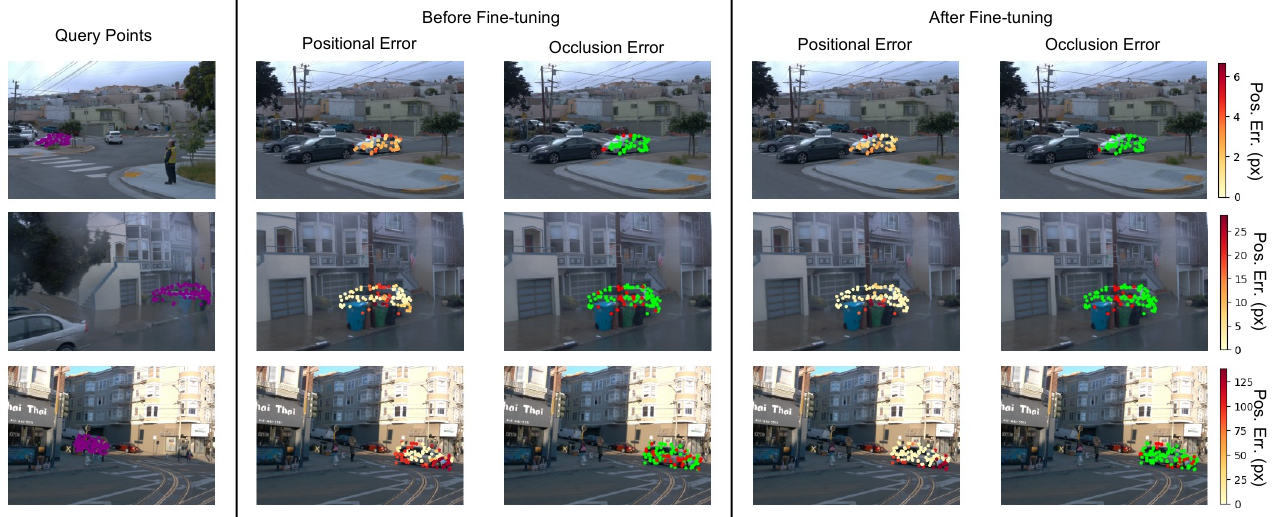}
    \vspace{-20pt}
    \caption{Positional tracking error (in pixels) and occlusion estimation error (binary) for \tapir~\cite{tapir} before and after fine-tuning on \name. Brighter colors indicate lower tracking error, and green dots on the occlusion map signify points that \tapir predicted correctly. Fine-tuning improves tracking accuracy especially around parts of the cars that blend with the background. Occlusion estimation also improves, albeit to a smaller degree.}
    \vspace{-10pt}
    \label{fig:finetune:drivetrack}
\end{figure*}

\subsection{Quantitative results}
\label{sec:finetune:quant}

Table~\ref{tab:finetuning} summarizes our results from fine-tuning. For each tracker, we report results on its reference models trained on a synthetic dataset and on our model fine-tuned with \name. We evaluate each model on \kubric, \davis, and \name.

\paragraphb{Improvements on \name.}
Fine-tuning improves AJ for \tapnet by 7\%, for \tapir by 5\%, and for \pipstwo by 4\%. The AJ of 84.1\% for a \tapir model fine-tuned on \name is on par (in terms of accuracy) with the equivalent for a synthetic benchmark, \eg~\tapir fine-trained on \kubric yields an AJ of 84.7\%. The largest improvement is in positional accuracy. As we noted in \S\ref{sec:motivation}, synthetic benchmarks do not model high frequency image imperfections, and fine-tuning on \name's annotations exposes the models to these artifacts. The improvement in OA is smaller, but fine-tuning helps nevertheless.

\paragraphb{Transferability.}
Fine-tuning on \name also transfers to \davis, which consists of real-world scenes drawn from a different distribution than the Waymo dataset~\cite{waymo} on which \name is built. In particular, for \tapnet, \tapir, and \pipstwo, AJ improves by 1-2\%. This shows that, even though \name is confined to datasets like autonomous driving videos due to its constraints on bounding box annotations and \lidar point clouds, trackers are capable of generalizing to real-world scenes more broadly. As expected, fine-tuning \tapir and \tapnet on \name degrade performance on the original training dataset, \kubric, which is synthetic.

\paragraphb{Depth completion.}
Table~\ref{tab:finetune-depth} compares AJ after fine-tuning \tapnet using the Nearest Neighbor and CompletionFormer~\cite{completionformer} depth completion methods that we explored in \S\ref{sec:data:depth}. While both models yield similar performance, we find that Nearest Neighbor outperforms CompletionFormer~\cite{completionformer} slightly on both \davis and \name. Although it can produce artifacts on the depth map, the max pooling that we apply on top of the Nearest Neighbor interpolation smooths out these artifacts. CompletionFormer ends up with more prediction errors that the max pooling feature cannot correct.

\subsection{Qualitative results}

\begin{figure}[t]
    \centering
    \includegraphics[scale=0.65]{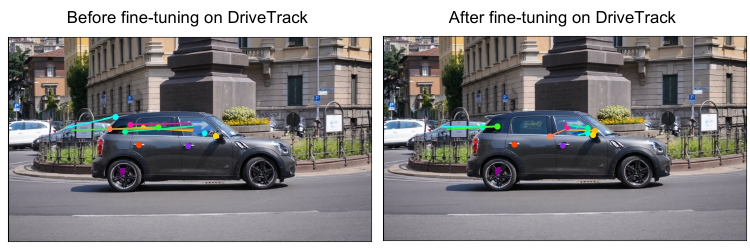}
    \vspace{-5pt}
    \caption{\tapir~\cite{tapir} fine-tuned on \name transfers well to scenes in \davis~\cite{davis}. Tracking improves, especially on the window.}
    \vspace{-10pt}
    \label{fig:finetune:tapir-davis}
\end{figure}

\paragraphb{\name.}
\fig~\ref{fig:finetune:drivetrack} compares \tapir's predictions before and after fine-tuning. We show points sampled on a query frame, as well as the positional error (in pixels) and occlusion estimation error (as a binary map) for each model. We show three representative scenes from \name. In all examples, the positional error generally decreases (\ie brighter colors) after fine-tuning, with the greatest improvement on points in the centers of the vehicles. Notice, for example, with the middle scene, that fine-tuning enables \tapir to track accurately despite lighting variations caused by rain. App.~\ref{app:finetuning} includes more annotated examples that illustrate improvements due to fine-tuning.

\paragraphb{\davis.}
\fig~\ref{fig:finetune:tapir-davis} shows \tapir before and after fine-tuning on a scene from \davis~\cite{davis}. Tracking on the window still exhibits significant error, but notice that fine-tuning yields a tangible improvement in positional accuracy. App.~\ref{app:finetuning} includes additional results on the \davis dataset.

\section{Sensitivity of trackers to keypoints}
\label{sec:sensitivity}

Despite the improvements from fine-tuning that we demonstrate in \S\ref{sec:finetune}, there is still an appreciable gap between the performance of these trackers on real-world scenes and the performance on synthetic datasets. \S\ref{sec:motivation} illustrates how trackers struggle particularly in scenes with complex lighting conditions and textures. While fine-tuning on \name can help make models more robust to these imperfections, we believe that precise tracking is fundamentally difficult in these settings. In this section, we use the scale of annotations in \name to quantify the sensitivity of keypoint trackers to particular keypoints.

\begin{figure}[t]
    \centering
    \includegraphics[scale=0.65]{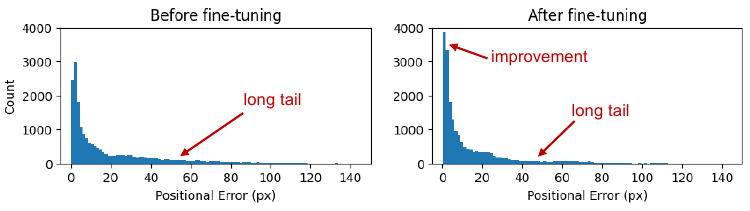}
    \vspace{-5pt}
    \caption{The distribution of tracking error across \name benchmark is heavy-tailed, even after fine-tuning. Certain kinds of keypoints are more susceptible to greater error.}
    \vspace{-5pt}
    \label{fig:sensitivity:hist}
\end{figure}

\subsection{Quantifying sensitivity with \name}
How is tracking error distributed over the \name benchmark? For each scene in the Waymo dataset~\cite{waymo}, we randomly sample 50 keypoints from a query frame and use \tapir~\cite{tapir} to track them over the entire video. We compute average positional tracking error for each keypoint against \name's ground-truth annotations, and consider the distribution of errors across all keypoints across all scenes. \fig~\ref{fig:sensitivity:hist} shows these histograms of \tapir's error before and after fine-tuning.

First, notice that fine-tuning reduces positional error noticeably on average: the fraction of points with tracking error less than 10 pixels is higher after fine-tuning. However, both distributions have long tails: the worst-case performance of \name improves marginally after fine-tuning. The 90th percentile error in both cases is over 60 pixels. This indicates that no amount of additional data will alleviate tail performance.

One might wonder why the tail performance of keypoint trackers matter: if we track tens of thousands of keypoints won't the effects at the tail be negligible? Invoking trackers like \tapir and PIPs++ is computationally-expensive, and is not practical for tracking thousands of pixels. Moreover, these trackers cater to the TAP formulation, where users want accurate tracks for very specific points. Those desired points might correspond to ones that yield error at the tail, and users currently have no way of knowing that.

\begin{figure}[t]
    \centering
    \includegraphics[scale=0.65]{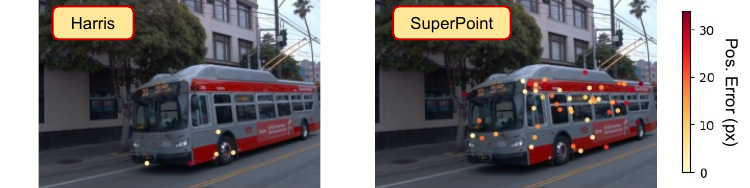}
    \vspace{-5pt}
    \caption{Existing keypoint selectors either offer a sparse set of options (Harris) or fail to avoid regions with visual imperfections (SuperPoint).}
    \vspace{-10pt}
    \label{fig:sensitivity:selector}
\end{figure}

\subsection{Characterizing keypoint sensitivity}
The annotations in \fig~\ref{fig:finetune:drivetrack} and \fig~\ref{fig:finetune:tapir-davis} indicate that points with high tracking error continue to correlate with visual imperfections, despite improvements from fine-tuning. For instance, notice from \fig~\ref{fig:finetune:drivetrack} that tracking performance is poorest on edges of the vehicle that are reflective (\eg top row, Prius windshield) and when the shadows partially cover the vehicle (\eg bottom row, minivan). In both examples, the lighting conditions vary between the query frame and the evaluation frame, and tracking error is exacerbated on reflective surfaces. Similarly, on the \davis dataset (\fig~\ref{fig:finetune:tapir-davis}), we find that \tapir continues to struggle on the car's windows. By contrast, \tapir tracks the door handle and wheel spokes perfectly.

\subsection{Towards track-aware keypoint selectors}
Given that positional tracking error correlates with visual imperfections and given the heavy-tailed error distributions, we argue that {\em keypoint selection} is as vital as keypoint tracking. We believe that users of any TAP-type tracker should also use a keypoint selector to strategically choose query points that will yield superior tracking accuracy with a high probability.

Keypoint selection from images has been an area of research in computer vision. The Harris Corner detector~\cite{harris} was one of the earliest methods to select trackable keypoints by looking at changes in image intensity to detect corners. SIFT~\cite{sift} is a featurization scheme that is popular for object recognition tasks. SuperPoint~\cite{superpoint} is a more recent method that predicts interest points by self-supervising on sets of images and various homographic transformations. While these techniques capture some elements of tracking tasks, none are well-suited for long-range tracking through occlusions and in cluttered scenes, which has become a popular domain for keypoint trackers~\cite{tapir, tapvid, pips, pointodyssey} over the last few years.

\fig~\ref{fig:sensitivity:selector} compares the positional accuracy achieved by \tapir for keypoints selected by Harris and by SuperPoint for an example in \name. Harris anchors well to corners and selects points that \tapir tracks accurately; however, it finds only 4 points to track in this scene. SuperPoint, by contrast, lists more candidates, but anchors to points on the window, which \tapir tracks poorly. Neither method is suitable for finding a large and robust set of trackable keypoints.

Effective keypoint selection will make trackers robust to visual imperfections. We need a keypoint selector that is aware that the downstream task is tracking. \name offers a rich dataset to train a feature representation to select robust keypoints in real-world scenes.

\section{Conclusion}
\label{sec:conclusion}

We developed \name, the first benchmark to automatically annotate long-range point tracks in real-world videos. We release the largest point tracking dataset on real-world scenes to date, consisting of 1 billion point tracks and 84 billion annotated points in total. \name's annotation workflow works with autonomous driving datasets consisting of point clouds, bounding box annotations, and camera poses. We develop a new way to robustly track a sequence of sparse point clouds that do not have 1:1 correspondence, and implement several refinements to be robust to labeling noise. Finally, we show that fine-tuning keypoint trackers with \name improves accuracy on real-world scenes, and we conduct a sensitivity analysis to motivate using keypoint selectors alongside trackers.

We believe that \name and the data generation workflow that we developed has several use cases outside of keypoint tracking, such as optical flow and structure-from-motion. Moreover, the ideas behind \name could be applied to other domains in the real-world with semi-annotated videos. For instance, we could consider building a similar workflow for annotate scenes imaged by iPhones and comparable smartphones, which can sense depth through \lidar and can record camera pose and orientation.

\clearpage
{
    \small
    \bibliographystyle{ieeenat_fullname}
    \bibliography{ref}
}

\clearpage
\appendix

\section{Implementation Details}
\label{app:impl}

We implement \name in Python using the Waymo Open Dataset SDK~\cite{waymo}. The dataset has separate tables for each data feature, such as video, bounding boxes, and \lidar data. Each feature table is further partitioned into Parquet files for each scene. We use Dask Distributed to merge the necessary tables one scene at a time before processing the annotations, using 8 workers with 2 threads each.

We preprocess the scene video, depth maps, and 3D \lidar point clouds before building the individual annotation point tracks. We use the Waymo SDK to convert the provided range images to 3D point clouds, and then  project the 3D point clouds to the image plane to generate depth maps (\S\ref{sec:data:depth}). For processing nearest-neighbor depth maps, we parallelize frames across CPUs. Using 32 CPU cores, processing 200 depth maps takes around 25 seconds. For depth maps generated using CompletionFormer~\cite{completionformer}, we spawn a process for each available GPU and use a multiprocessing queue to pass jobs to each process. On 6 NVIDIA V100 GPUs, processing 200 depth maps takes around 35 seconds.

After preprocessing the scene and caching the videos, depth maps, and 3D point clouds, we process each annotation (object) in parallel as described in (\app\ref{sec:data}). In total using 32 CPU cores, each annotation takes about 10 minutes to complete.

\section{\name Annotations}
\label{app:annotations}

\begin{figure*}[t]
    \centering
    \includegraphics[width=\textwidth]{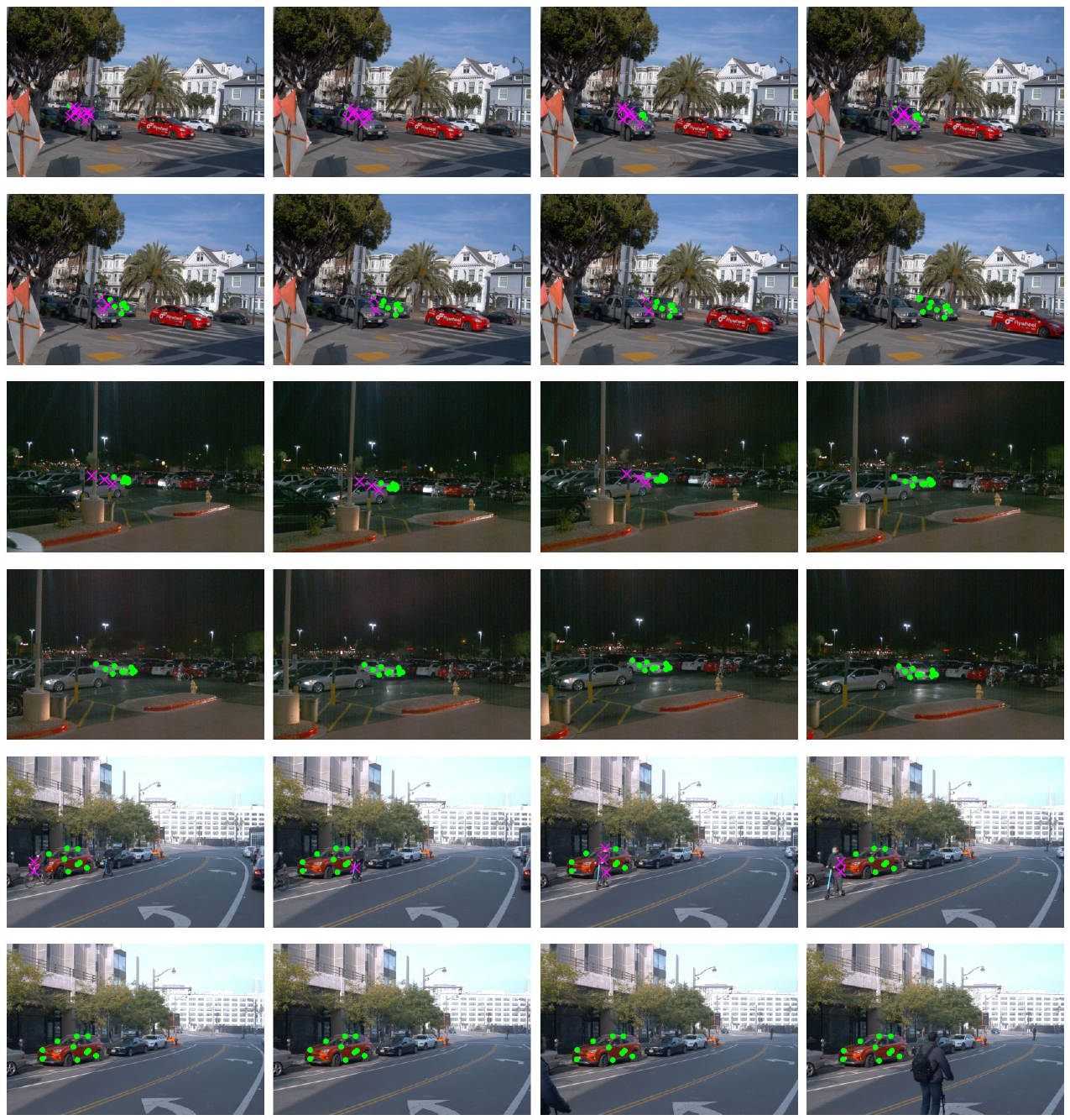}
    \caption{Example annotations from \name on three scenes, spanning a variety of lighting and weather conditions. Each batch of two rows corresponds to a new scene, spaced four frames apart on a 30-frame subsection of the scene video.}
    \label{fig:drivetrack-examples}
\end{figure*}

\begin{figure*}[t]
    \centering
    \includegraphics[scale=0.27]{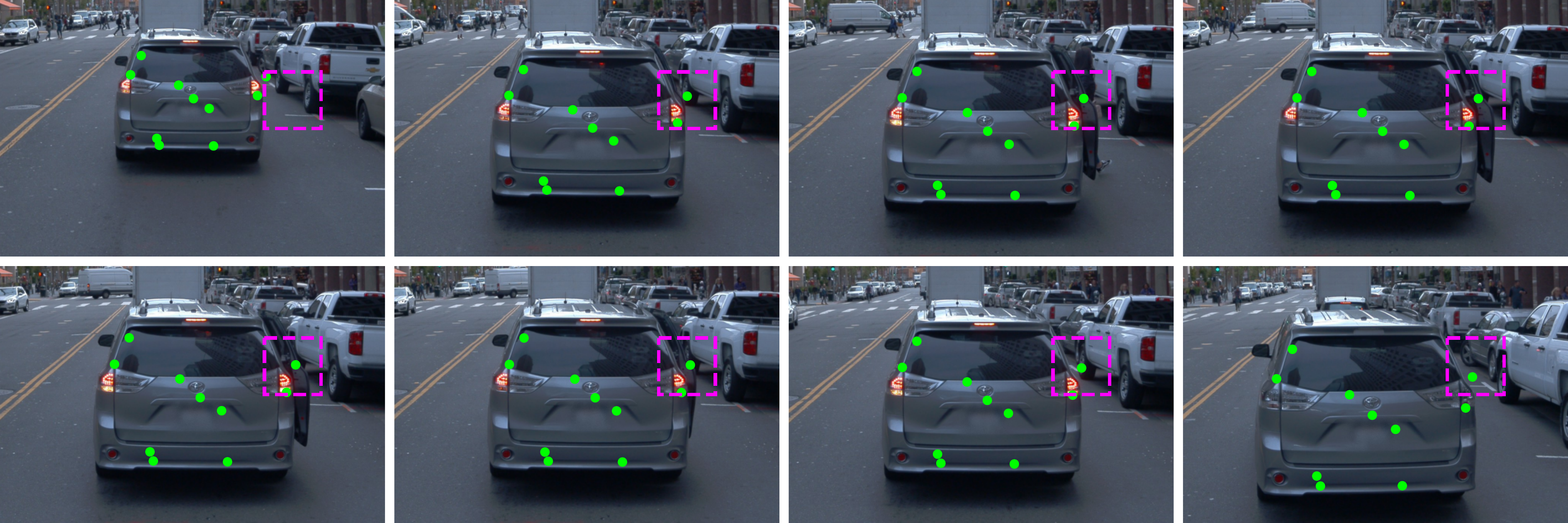}
    \caption{Example of a failure in \name's annotations, highlighted by the magenta box. When the van door closes, the point selected on the door frame becomes dangled in space instead of tracking the door as it closes.}
    \label{fig:limitation}
\end{figure*}

\subsection{Video animation}
We recommend watching the supplementary video to see several qualitative visualizations of \name's annotations. Our video includes 11 scenes, covering a variety of lighting conditions, weather patterns, and occlusion configurations. For each scene, we show 10 randomly-sampled points from \name's annotations; each point disappears whenever \name labels it as occluded.

\subsection{Examples}

\fig~\ref{fig:drivetrack-examples} visualizes the annotations computed by \name for several scenes in the Waymo dataset~\cite{waymo}. The top two rows illustrate how \name robustly detects different types of occlusions, from both cars and signposts. The middle two rows demonstrate a nighttime scene with vehicle occlusions. The bottom two rows demonstrate occlusions from pedestrians in the scene.

\subsection{Limitations}

While most of \name's annotations are high-quality, there are a few edge cases where \name fails. \fig~\ref{fig:limitation} illustrates one example, where \name fails to track a point on a sliding door on a parked van. Due to the rigid body assumption, \name selects points on the open door and then tracks them through all frames. However, as the door closes, the points should follow the door and eventually become occluded once the door closes. Instead, \name leaves such points dangling in open space. 

A simple way to mitigate these errors is to filter out points whose speeds deviate significantly from the annotated ones (\S\ref{sec:data:robustness}). A more robust method is to estimate surface normals from the 3D point cloud, which would help better track the depth contours of the vehicle. We leave this to future work.

\section{Additional Fine-tuning Results}
\label{app:finetuning}

\begin{figure*}[t]
    \centering
    \includegraphics[width=0.98\textwidth]{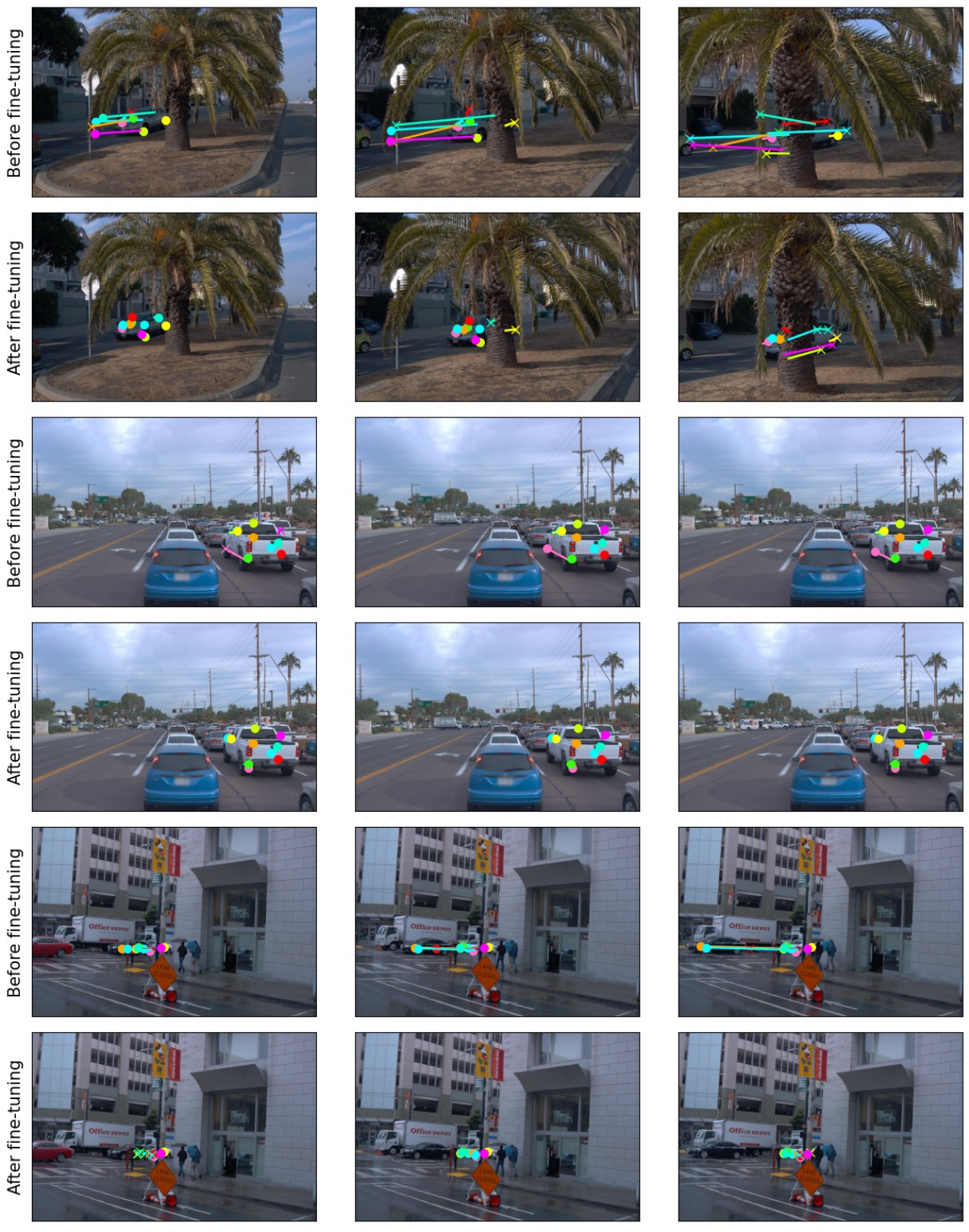}
    \vspace{-12pt}
    \caption{Point tracks predicted by \tapir before and after fine-tuning. The markers indicate the locations predicted by each model, and the line segments lead to their respective ground-truth locations. $\bullet$ denotes points predicted as visible, and $\times$ denotes points predicted as occluded.
    }
    \vspace{-10pt}
    \label{fig:app:finetune1}
\end{figure*}

\begin{figure}[t]
    \centering
    \includegraphics[scale=0.65]{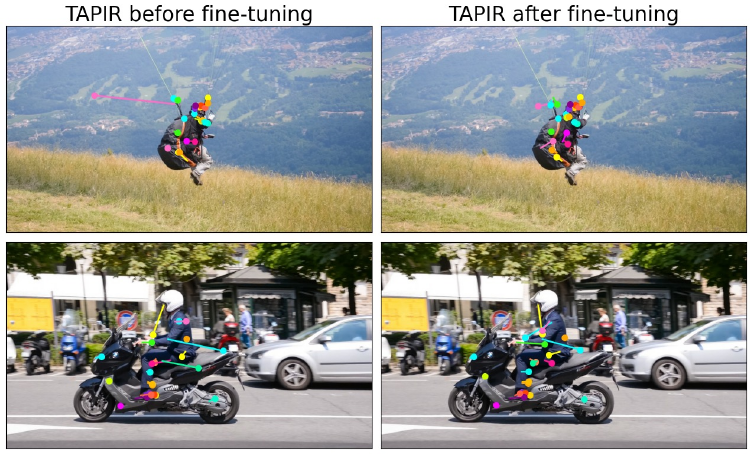}
    \caption{Point tracks predicted by \tapir on scenes from \davis.}
    \vspace{-10pt}
    \label{fig:app:finetune-davis}
\end{figure}

\fig~\ref{fig:app:finetune1} shows additional results from fine-tuning \tapir on \name. Each row shows three frames from a scene in \name, and visualizes tracking performance before and after fine-tuning. Fine-tuning on \name consistently improves tracking performance.

\fig~\ref{fig:app:finetune-davis} shows the transfer potential of \tapir on more scenes from the \davis dataset. We find that fine-tuning in particular improves the tracking accuracy of outliers.


\end{document}